\documentclass{elsarticle}
\usepackage{amsmath}
\usepackage{amssymb}
\usepackage{graphicx}
\usepackage{multirow}
\usepackage{ragged2e}
\usepackage{xcolor}
\usepackage{comment}
\usepackage{enumerate}
\usepackage{url}
\usepackage{tikz}
\usepackage{array}
\newcolumntype{?}{!{\vrule width 1.1pt}}
\begin{document}

\title{Bridging the Applicator Gap with Data-Doping: Dual-Domain Learning for Precise Bladder Segmentation in CT-Guided Brachytherapy}
\author[add1]{Suresh Das}
\ead{dassureshster@gmail.com }
\author[add3]{Siladittya Manna}
\ead{smanna@hkbu.edu.hk}
\author[add6]{Sayantari Ghosh$^*$}
\ead{sayantari.ghosh@phy.nitdgp.ac.in}

\address[add1]{Narayana Superspeciality Hospital, 120/1 Andul Road, Howrah, India }

\address[add3]{Department of Computational and Data Sciences, Indian Institute of Science, Bengaluru, India}
\address[add6]{Department of Physics, National Institute of Technology Durgapur, India}

\cortext[cor1]{Corresponding author}

\begin{abstract}
Performance degradation in deep models caused by covariate shift is a well-known challenge. This, in turn, raises the open question of whether samples from one distribution can aid learning when combined with samples from a shifted distribution. We explore this issue in the context of  bladder segmentation  in CT-guided gynecological brachytherapy, which is essential for optimizing dose delivery and minimizing organ-at-risk exposure. As datasets containing applicator-inserted scans are extremely rare, automation by the development of deep segmentation models remains difficult. Although many gynecological CT-scans, performed for other clinical purposes (i.e., without a brachy-applicator inserted) are available, their potential value for organ segmentation in applicator-inserted images remains questionable and largely unexplored, due to the substantial anatomical deformation and imaging artifacts caused by the applicator. This study introduces a dual-domain learning approach that integrates data from both applicator-present (With applicator: \textbf{WA}) and applicator-absent (no applicator: \textbf{NA}) CT-scans to enhance segmentation robustness and generalizability. Curating a comprehensive assorted dataset, we investigate this issue to find the following: while \textbf{NA}-data alone cannot adequately capture the anatomical and artifact-related characteristics of \textbf{WA} images, introducing a modest percentage of \textbf{WA} data into the training set yield significantly improved results, greatly boosting the performance metrics. It is non-trivial to understand the minimum amount of \textbf{WA} images required in the training set to make the model robust to the covariate shift. We further perform systematic experiments by data augmentation with this anatomically similar and abundantly available \textbf{NA}-data to optimize  the variability in anatomy and imaging. Experimental results using multiple deep learning architectures in the axial, coronal, and sagittal planes demonstrated that a mere $10-30\%$  \textbf{WA} \textit{data doping} in otherwise predominant \textbf{NA} training produces substantial gains in segmentation accuracy and achieves performance on par with models trained exclusively on \textbf{WA} data. The proposed dual-domain strategy achieved DSC of up to 0.94 and IoU of up to 0.92, indicating effective domain adaptation and improved clinical reliability, to address the unavailability of applicator-inserted datasets. This work underscores the value of integrating \textit{assorted datasets} for deep learning-based segmentation in brachytherapy to overcome data scarcity and to enhance treatment planning precision.
\end{abstract}
\begin{keyword}
Bladder segmentation \sep
CT-guided brachytherapy \sep
Dual-domain learning \sep
Data doping
\end{keyword}

\maketitle

\section{Introduction}
\noindent Cervical cancer is the third most prevalent cancer among women worldwide, accounting for nearly 530,000 new cases and about 250,000 deaths annually \cite{parkin2005global}. While its incidence in developed nations has declined 
over the past five decades, largely due to the widespread implementation of improved cervical cytology screening and human papillomavirus (HPV) vaccination programs \cite{gustafsson1997international, koutsky2002controlled}, in low-income countries, it remains the second leading cause of cancer-related mortality in women \cite{kamangar2006patterns}. 
Brachytherapy, a form of internal radiation therapy, involves placing radioactive sources directly into or near a tumor, allowing high radiation doses to reach the target while sparing the surrounding healthy tissue \cite{otter2019modern,major2022value}. In brachytherapy, organ segmentation is essential because it allows precise identification of organs-at-risk (OARs) and target volumes, guaranteeing the best possible dose delivery to the tumor while reducing radiation exposure to nearby healthy tissues \cite{tanderup2016effect}. Automated or partially automated segmentation enhances the efficiency of treatment planning and minimizes variability among observers when compared to traditional manual contouring. Accurate segmentation of organs ultimately boosts treatment precision and patient safety in image-guided brachytherapy \cite{rhee2021automation,yan2021status}. Segmentation is a crucial task in brachytherapy as it involves accurately delineating organs-at-risk (OARs), target volumes, and applicators from medical images to ensure precise dose delivery. Accurate segmentation enables optimized treatment planning by minimizing radiation exposure to surrounding healthy tissues while achieving adequate tumor coverage \cite{siddique2020artificial}. The precision of segmentation plays a crucial role in determining dose distribution, treatment results, and ensuring patient safety in image-guided brachytherapy \cite{kalantar2021automatic,eustace2024current}.

Manual segmentation can be labor-intensive and susceptible to differences between observers, which may affect the uniformity and quality of treatment. For this reason, there is a growing trend towards adopting automated or semi-automated segmentation techniques that utilize deep learning to improve efficiency and consistency. Thanks to deep learning, a branch of artificial intelligence built on multi-layered neural networks, computational models can automatically discover intricate patterns from massive datasets without the need for explicit feature building. \cite{lecun2015deep,chowa2025low}. In medical physics, it is increasingly applied to  segmentation tasks, improving both accuracy and efficiency.  

Datasets are essential for the segmentation of OARs in brachytherapy, as they provide the foundation for training, validating, and assessing deep learning models that support the automated identification of OARs and target regions, thereby enhancing both reliability and efficiency in treatment planning. However, datasets of brachytherapy scans for organ segmentation are often scarce, which restricts the ability of deep models to learn robust anatomical representations. Training with limited data restricts a model’s ability to learn meaningful patterns, often resulting in underfitting. As a consequence, the model fails to capture the underlying data distribution and generalize effectively.

In case of data scarcity, these models generally turn to different data augmentation strategies to improve model performance. 
There are three kinds of common augmentation strategies: affine image
transformations, elastic transformations and various approaches for generating artificial data.
All these methods have their own limitations. Affine methods produce very correlated
images, and therefore can offer very little improvements for the deep-network training
process and future generalization over the unseen test data \cite{nalepa2019data}. They can also generate
anatomically incorrect examples \cite{das2024bladder}. Data augmentation algorithms based on unconstrained
elastic transformations of training examples can introduce shape variations. They can bring
lots of noise and damage into the training set if the deformation field is varied, producing a
totally unrealistic synthetic MRI scan of a human brain \cite{mok2018learning}. If the simulated tumors were placed
in \textit{unrealistic} positions, it would likely force the segmentation engine to become invariant
to contextual information and rather focus on the lesion’s appearance features \cite{dvornik2018modeling}. Generative models, though have shown prospects with successful domain adaptation  \cite{zhu2017unpaired}, still need huge amount of data to train the generative models. 

Standardized, expert-annotated datasets have been fundamental to progress in medical image segmentation by enabling reproducible training, fair comparison, and rigorous evaluation. Benchmarks such as MSD and BraTS demonstrated the impact of multi-task, multi-modality and large-scale expert annotations on developing robust and generalizable methods \cite{antonelli2022medical,adewole2023brain}. Organ- and abdomen-focused resources, including KiTS and AMOS, highlighted the importance of high-quality voxel-level labels and multi-center variability for reliable generalization \cite{uhm2023exploring,ji2022amos}. More recent unified benchmarks like MedSegBench revealed limitations of single-task training and facilitated systematic cross-dataset evaluation \cite{kucs2024medsegbench}. 

\section*{Motivation}
\begin{figure}[!htbp]
\begin{center}
    \includegraphics[width=\linewidth]{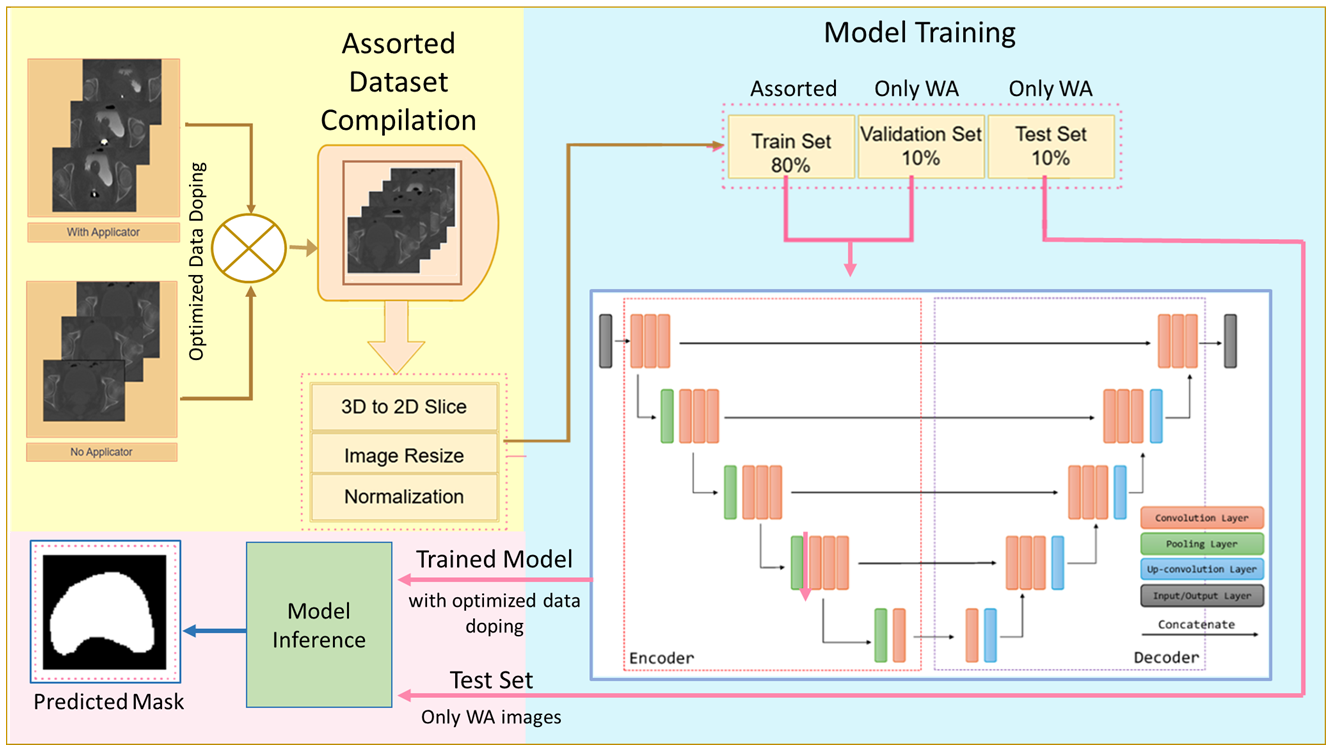}
    \caption{Block diagram of the process flow for model training and validation with optimally assorted dataset. \textbf{WA} denotes With-applicator images.}
    \label{Schematic}
\end{center}
\end{figure}

In this work, to address the data scarcity issue in gynecological brachytherapy, we adopt a novel approach. While there are very few data available usually with brachy-applicator
inserted in a patient (\textbf{WA} data), there are several scans without the applicator present (\textbf{NA} data) are easily available; these scans are typically performed for other routine purposes. It requires an extensive exploration of how the shift in input data distribution from $P(X_{NA})$ to $P(X_{WA})$ affects the performance of a model in a downstream task. Here, $X_{NA}$ and $X_{WA}$ are image samples without or with an applicator, respectively. In this study, we explore the bladder segmentation task, and thus, the distribution of the target label $P(y|X_i)$, $i\in\{WA,NA\}$, remains unchanged irrespective of the presence or the absence of the applicator. We also wanted to study if samples $x \sim P(X_{NA})$  can be of any use for
augmenting and amplifying the model performance when \textbf{WA} data is scarce (block diagram shown in Fig. \ref{Schematic}). This is not a
trivial problem, as the presence of the applicator distorts the organs significantly and also distorts
images to a certain extent, introducing imaging artifacts (shown in Fig. \ref{Distorted image}) and making the value of \textbf{NA} data questionable in this context. On the other hand, due to similar contextual information,
organization, and domain of the images, this might enhance the segmentation task. In some recent works, generated synthetic data was proportionally mixed with real data to enhance the
model performance \cite{liang2024segmentation}. We plan to take a similar strategy, but not with synthetic data, but
anatomically similar data of abundant availability. Thus, our goal is to see if a client has $n$ no. of \textbf{NA} images and and $m$ no. of \textbf{WA} images in the training set (where $m$ might be insufficient in number for model training, but, \(m\ge 0\) and \(n \gg m\)), can the training process benefit from the presence of these \textbf{NA} images, or not. We thus intend to
address the following research questions:
\begin{enumerate}
\item For automated organ-at-risk (bladder) segmentation, can a model be trained solely
using \textbf{NA} data, to detect the organ of interest efficiently for a \textbf{WA} test dataset?
\item Keeping the test set fixed, can the performance be improved if an assorted
dataset is used instead, where both \textbf{WA} and \textbf{NA} data will be present in training set, in certain proportions?
\item Is there any optimal percentage of mixing, that gives a comparable performance,
at par with the best performing model when trained exclusively on
proportionally large \textbf{WA} dataset?
\end{enumerate}


Thus, to be more precise, our motivation is as follows: with $n$ no. of \textbf{NA} images forming a set $\mathcal{C}_{NA}$ such that $\mathcal{C}_{NA}=\{x|x \sim P(X_{NA})\}$ and $|\mathcal{C}_{NA}|=n$, and $m$ no. of \textbf{WA} images forming a set $\mathcal{C}_{WA}$ such that $\mathcal{C}_{WA}=\{x|x \sim P(X_{WA})\}$ and $|\mathcal{C}_{WA}|=m$ in the training set (where $m$ might be insufficient in number for model training, but, \(m\ge 0\) and \(n \gg m\)), we intend to study the effect of presence of the set $\mathcal{C}_{NA}$. For automated organ-at-risk (bladder) segmentation, thus we first look into a deep model trained using solely $x_{train} \sim P(X_{NA})$, to detect the organ of interest efficiently in test samples $x_{test} \sim P(X_{WA})$, and investigate the findings. Next, keeping the test set $\mathcal{D}_{test}=\{x_{test}|x_{test} \sim P(X_{WA})\}$ fixed, we look for improvement of performance, in any, using an assorted training set $\mathcal{C}_{a}=\mathcal{C}_{NA}\cup \mathcal{C}_{WA}$ for training, with a goal of finding an optimal ratio $n:m$, for various different deep models. To address these issues, we perform a series of experiments where a standard U-Net model is
trained several times with differently assorted training sets, and the performance has been
evaluated for the same test set of \textbf{WA} data. We refer to this data mixing as \textit{data-doping}, drawing analogy from Semiconductor Physics, where a small percentage of a foreign substance is added, causing a drastic improvement in the performance. We find an optimum percentage of data-doping repeating this experiment systematically, and further generalize it for other established segmentation models, like U-Net++, Half-UNet, RRDB U-Net,  DC-UNet, Attention U-Net etc. The rest of the paper is organized as follows: in Section 2, we discuss the methodology in detail; in Section 3, we consolidate all important findings, and we conclude with a brief discussion in Section 4.

\section{Material and methods}
\begin{figure}[!htbp]
\begin{center}
    \includegraphics[width=0.8\linewidth]{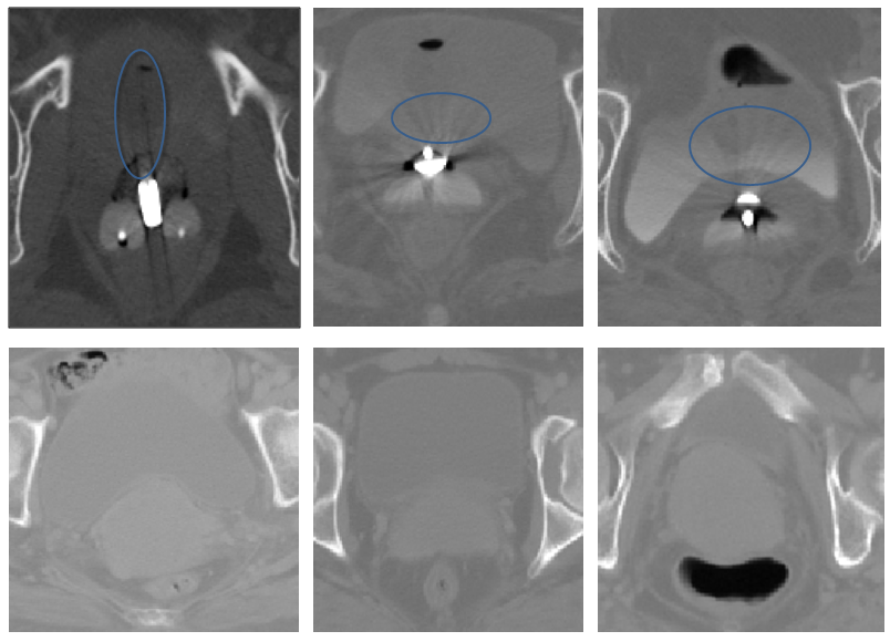}
    \caption{{Depiction of distorted image due to the presence of the brachy-applicator. Top Panel: Artifacts due to applicator insertion have been marked. Bottom Panel: No such artifact.}}
    \label{Distorted image}
\end{center}
\end{figure}

This study focuses on the development and evaluation of deep learning models designed for automated bladder segmentation in brachytherapy planning, using a medical imaging dataset of general CT scans, acquired for other purposes than brachytherapy. We aim to create a workflow for utilizing this \textbf{NA} dataset with the goal of bladder segmentation in \textbf{WA} scans. {This research suggests a method that merges a limited amount of \textbf{WA} with a more substantial amount of \textbf{NA} to meet the required sample size and variability for testing.} This strategy successfully lowers research expenses while maintaining the quality and dependability of the experimental results. In gynecological brachytherapy, placing an intracavitary applicator in the vagina can cause considerable distortion in CT scan images, which may result in a deceptive representation of the bladder. The applicator creates a strong shadow or artifact on the image, which disrupts the normal appearance of the bladder. Because of this, the bladder wall may look pushed, bent, or changed, even though it is actually normal. This distortion is not merely an imaging effect, which makes it more challenging to accurately draw the bladder during treatment planning. As a result, there is an increased risk of inflating the estimated bladder volume and incorrectly assessing its spatial relationship to radiation dose distribution, as illustrated in Fig. \ref{Distorted image}.

\subsection{Dataset establishment}
Developing a reliable dataset for bladder segmentation in brachytherapy requires careful consideration of anatomical differences, the type of imaging used, and whether the images include or exclude applicators. In this research, a diverse dataset has been assembled, incorporating cases both with and without intracavitary applicators, to facilitate model generalization across various clinical contexts. We considered a freely available raw dataset \cite{das2025residual} for mask creation, where CT-guided intracavitary brachytherapy treatment planning was under consideration for a portion of patients (i.e., \textbf{WA} data). For rest of the patients the scans were performed for routine non-brachy purposes (i.e., \textbf{NA} data). We choose this dataset as this perfectly addresses challenges in bladder deformation due to applicator placement \cite{chi2022clinical}, differences in contrast between imaging modalities \cite{potter2007clinical}, and inter-observer variability \cite{hellebust2013dosimetric}. On this data, manual annotations were performed by expert radiation oncologist following established contouring guidelines \cite{viswanathan2014comparison}, with quality assurance by a senior radiation oncologist. Standardization protocols were also employed for voxel spacing and intensity normalization \cite{isensee2021nnu}. Including cases \textbf{WA} allows for the evaluation of segmentation algorithms in high-dose gradient regions, critical for accurate dose planning \cite{dimopoulos2012recommendations}. This comprehensive assorted dataset supports our purpose of required domain adaptation, given the variability introduced by applicator's presence or absence \cite{kamnitsas2017unsupervised}. By including a wide range of anatomical differences and procedural conditions, the dataset supports the development of deep learning models that can achieve robust segmentation performance in real brachytherapy settings. Within the training dataset, \textbf{NA} and \textbf{WA} instances are randomly sampled respectively, post data augmentation, in accordance with specific proportions.

\subsection{Model selection}

In this study, multiple deep learning architectures were employed for automatic bladder segmentation using an assorted dataset comprising varied patient scans from CT scans. The dataset was preprocessed to ensure uniform resolution, intensity normalization, and organ-specific cropping to enhance model focus, and extensive experiments were carried out using U-Net architecture to evaluate the optimal mixing percentages. Next, five more segmentation models were implemented (i.e., U-Net++, Half-UNet, DCNet, Attention U-Net, and RRDB U-Net), to validate the generalization of the findings. Out of the used models, U-Net and U-Net++ are encoder-decoder frameworks with skip connections that facilitate gradient flow and multi-scale feature extraction \cite{ronneberger2015u,zhou2018unet++,harrison2023state}. U-Net++ is an enhanced version of U-Net that introduces redesigned skip connections and nested convolutional blocks to reduce the semantic gap between encoder and decoder features, improving segmentation accuracy for multiorgan detection \cite{zhou2018unet++,tejashwini2025enhanced}. Half-UNet reduces computational complexity by employing a half-channel encoder-decoder pathway \cite{zhang2021u}. DCNet leverages dense connections and context modules to retain fine-grained spatial information \cite{chen2023medical}. Attention U-Net integrates attention gates and has been used before to emphasize relevant bladder regions and lung edge detection \cite{oktay2018attention,li2025ae}, while RRDB U-Net, inspired by ESRGAN, incorporates Residual-in-Residual Dense Blocks for rich hierarchical features and stability \cite{wang2018esrgan}. All models were trained using training and validation splits with data augmentation techniques to improve generalizability. Training was conducted using Python with the Adam optimizer and early stopping to prevent overfitting.

\subsection{Evaluation Indicator}

In brachytherapy-based bladder segmentation, particularly using assorted datasets containing images \textbf{NA} and \textbf{WA}, robust evaluation metrics are crucial for validating model performance. The Dice Similarity Coefficient (DSC) and Intersection over Union (IoU) are the most commonly employed indicators for assessing segmentation quality due to their spatial overlap quantification capability.
The DSC defined as
\begin{equation}
\text{DSC} = \frac{2|A \cap B|}{|A| + |B|},
\end{equation}
which evaluates the degree of overlap between the predicted and ground truth masks, effectively handling small organ sizes such as the bladder.
IoU, or the Jaccard index, calculated as 
\begin{equation}
\text{IoU} = \frac{|A \cap B|}{|A \cup B|},
\end{equation}
which provides a stricter measure of similarity by considering both false positives and false negatives.
Assorted datasets, comprising images \textbf{WA} (e.g. tandem-ovoid) and \textbf{NA}, introduce anatomical and intensity variability. Evaluation metrics must reflect robustness against such heterogeneity. Several studies have shown DSC to be more stable in cases of high class imbalance, such as with small bladder volumes or obscured anatomy due to applicators \cite{fu2021review,coroamua2023fully}. Conversely, IoU provides a better distinction in low-overlap scenarios \cite{xu2019efficient}. Researchers recommend using both metrics in tandem for comprehensive evaluation \cite{isensee2021nnu}. 
\section{Results}
\label{section:Results}
To establish this work, all computational experiments were conducted on a workstation equipped with an NVIDIA RTX A4000 GPU, Intel (R) Xeon (R) processor, and 64 GB RAM. The models were implemented in Python 3.10.12 using TensorFlow 2.10.0 and Keras frameworks. Training and testing pipelines were automated for reproducibility, and random seeds were fixed to ensure consistent results. Model checkpoints and logs were maintained for all experiments. For experiments on all three planes, Axial, Sagittal and Coronal, we used a batch size of 32, and each model was trained for 50 epochs or till validation loss plateaued to prevent over-fitting. We also transformed the input slices to the dimensions $128 \times 128$ before it is processed by the model. 

In this work, the result analysis is conducted to establish proper justification of mixing of the \textbf{NA} and \textbf{WA} for utilizing the U-Net architecture. The detail investigations consider the 2D image slices of the axial, coronal and sagittal planes separately from recent clinical CT image set \cite{das2025residual} comprising 20 patients \textbf{NA} and 20 patients \textbf{WA}. For the comparative analysis, six deep-learning models are considered, e.g U-Net, Half-UNet, RRDB U-Net, U-Net++, DC-UNet and Attention U-Net. In order to perform this work, the He- initializer is utilized with Adam optimization, 32 batch size with 50 epochs for the investigation in every plane. The dice loss is considered to be the loss function. 

\begin{figure}[!htbp]
\begin{center}
    \includegraphics[width=0.8\linewidth]{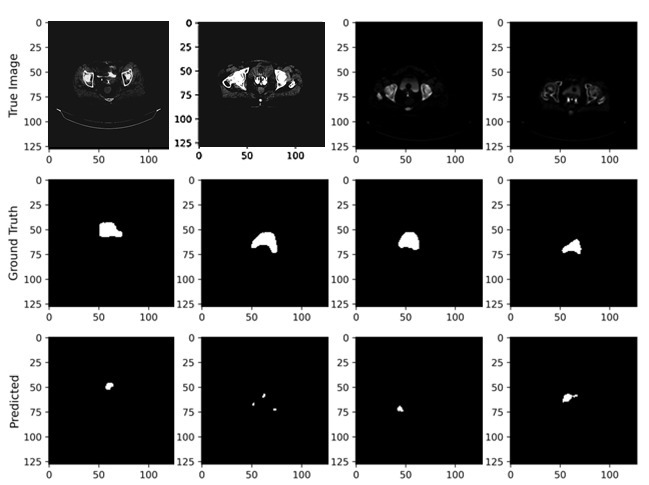}
    \caption{Failure of the training in bladder segmentation  with exclusive NA data training. The top, middle and bottom panels represent original CT image for Axial plane, corresponding ground truth masks and segmentation results obtained for test dataset (\textbf{WA} data) respectively.}
    \label{NAtraining}
\end{center}
\end{figure}

\subsection{Data selection}
As the final goal is to improve brachytherapy treatment planning, the test dataset was fixed throughout the study and comprised of only \textbf{WA} data samples. However, variability of training dataset is at the heart of this study. To ascertain the importance and ideal proportion of \textbf{NA}-data within the training dataset, initial experiments were performed exclusively on a dataset compromising solely with \textbf{NA}, serving as a control condition for this investigation. Following this, trials were undertaken using a blend of \textbf{NA} and \textbf{WA} in varying ratios to ascertain the optimal fusion of the two. 
This iterative process was aimed at identifying the most effective data-doping for training, and ensuring the optimum results for bladder segmentation.The proportionally doped training dataset was also used for generalizing the framework for different deep learning models. Key findings of these experiments are discussed in the upcoming sections.

\subsection{Training only with \textbf{NA} data}
Our first research question was following: for automated organ-at-risk (bladder) segmentation, can we train a model solely using \textbf{NA} data, to detect the organ of interest efficiently in \textbf{WA} test dataset? This we begin our initial experiments where the training data comprised of only \textbf{NA} samples. The deep model is chosen as U-Net, to establish the methodology. As mentioned before, test dataset was fixed throughout all experiments and comprised of only \textbf{WA} samples. The segmentation task fails critically showing an average IOU $~0.71$. Visual results have been shown in Fig. \ref{NAtraining} that depicts come extreme failure cases, especially in the axial plane where the IOU is close to $0.62$. These results show that only \textbf{NA} training in incapable of performing accurate segmentation, and reason could be lying in the requirement of domain adaptation.

 \begin{table}[!htbp]
\centering
\normalsize
\caption{{\normalsize Analysis of the effect of data doping on the segmentation performance.}}
\label{tab:mixing}
\begin{tabular}{|c|c|c|c|c|c|c|}
\hline
{\scriptsize Training Set} & \multicolumn{3}{c|}{IOU }& \multicolumn{3}{c|}{Dice } \\ 
\cline{2-7}
 {\scriptsize Ratio (NA:WA)} & Sagittal & Axial  & Coronal & Sagittal & Axial  & Coronal \\ \hline
01:09            & 0.9225   & 0.8758 & 0.9164 & 0.9442 & 0.8959 & 0.9324 \\ \hline
02:08           & 0.9165   & 0.8875 & 0.9146 & 0.9442 & 0.9082 & 0.9307 \\ \hline
03:07           & 0.919   & 0.8777 & 0.9137 & 0.9408 & 0.8986 & 0.93 \\ \hline
04:06     & 0.9197   & 0.8737 & 0.9119 & 0.9413 & 0.8952 & 0.9239 \\ \hline 
05:05     & 0.9156   & 0.8899 & 0.9037 & 0.9384 & 0.9113 & 0.9201 \\ \hline 
06:04     & 0.9109   & 0.8650 & 0.9029 & 0.933 & 0.8850 & 0.9206 \\ \hline 
07:03     & 0.8831   & 0.8899 & 0.9013 & 0.9066 & 0.9113 & 0.9193 \\ \hline
08:02     & 0.8636   & 0.8575 & 0.8778 & 0.8636 & 0.8807 & 0.8917 \\ \hline 
09:01     & 0.8313   & 0.7847 & 0.864 & 0.8504 & 0.802 & 0.8786 \\ \hline 
 Only WA     & 0.9202   & 0.9015 & 0.916 & 0.9411 & 0.9213 & 0.924 \\ \hline 
 Only NA     & 0.7665   & 0.6237 & 0.796 & 0.7672 & 0.6303 & 0.7986 \\ \hline 

 \end{tabular}
\end{table}
\subsection{Data doping: Investigation with Assorted Data}
To incorporate dual domain adaptation, we proceed with data doping, i.e., mixing some \textbf{WA} data in the training set. This will address our second research question, i.e., keeping the test set fixed, can the performance be improved if an assorted dataset is used instead, where both \textbf{WA} and \textbf{NA} data will be present in training set in certain proportion. We begin with a training set made of \textbf{WA} data solely, to set the target IOU for present scenario. Our goals for the data doping experiments were to achieve a value close to that, utilizing the \textbf{NA} data, as much as possible. We look for an optimum doping where:
\begin{equation}
   \frac{IOU_{WA}-IOU_{doped}}{IOU_{WA}}\leq0.05
   \label{criteria}
\end{equation}

\begin{figure}[!htbp]
   \centering
    \includegraphics[width=0.9\linewidth]{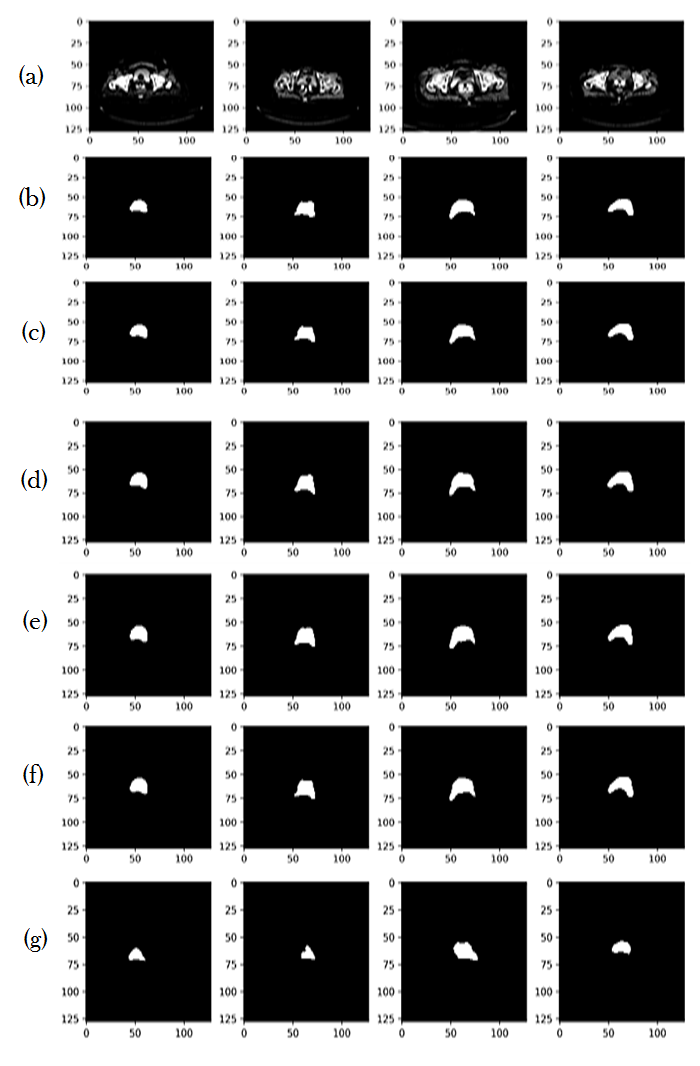}
    \caption{Bladder segmentation in Axial plane with different proportional data doping. The panels (a) and (b) represent original CT image and corresponding ground truth masks respectively. Panels (c)-(g) represent segmentation results for different \textbf{NA:WA} ratio in training data: (c) 1:9, (d) 3:7, (e) 5:5, (f) 7:3, (g) 9:1. }
    \label{Axialimage}
\end{figure}

We observe a drastic average improvement of $10.6\%$ in IOU from the \textbf{NA} training experiments, as soon as a mere 10\% \textbf{WA} data is added to the training data, keeping the sample size fixed. The detailed simulation results are obtained using the U-Net architecture in different doping proportions for all the three planes have been compiled in Table \ref{tab:mixing}. The visual results for these finding have been reported in Fig. \ref{Axialimage}. Examination of the values reveal that the mixing criteria is \textbf{NA:WA}$=07: 03$ gives an enormous improvement and satisfy criteria in Eq. \ref{criteria}  for all three planes in case of the U-Net architecture. This is a powerful result that shows a minor 30\% doping can make the model perform at par, even when the \textbf{WA} data is barely available. We proceed for the third set of experiments, to evaluate the generalizability of this framework for other deep models considering this optimum doping level in training set.

\begin{table}[!htbp]
\centering
\caption{\normalsize Comparison of IOU and DICE scores for different models across planes. Metrics have been reported for only \textbf{NA} training and \textbf{NA:WA}$=7:3$ data doping. The improvements can be noted in bold fonts. }

\label{tab:model_results}
\normalsize
\begin{tabular}{|>{\centering}p{1.2cm}|>{\centering}p{1.2cm}|>{\centering}p{1cm}p{1cm}p{1.1cm}|>{\centering}p{1cm}p{1cm}p{1.1cm}|}
\hline
\multirow{2}{*}{Model} &
Training&
\multicolumn{3}{c|}{IOU} &
\multicolumn{3}{c|}{DICE} \\
\cline{3-8}
 & Set & Sagittal & Axial & Coronal & Sagittal & Axial & Coronal \\
\hline

{\scriptsize U-Net} & \scriptsize{Only NA} & 0.7665 & 0.6237 & 0.7960 & 0.7672 & 0.6303 & 0.7986 \\
     & \textbf{07:03} & \textbf{0.8831} & \textbf{0.8617} & \textbf{0.9013} & \textbf{0.9066} & \textbf{0.8829} & \textbf{0.9193} \\
\hline

{\scriptsize U-Net++} & \scriptsize{Only NA} & 0.7789 & 0.7332 & 0.8082 & 0.7790 & 0.7377 & 0.8088 \\
   & \textbf{07:03} & \textbf{0.8553} & \textbf{0.7996} & \textbf{0.8805} & \textbf{0.8782} & \textbf{0.8209} & \textbf{0.8976} \\
\hline

{\scriptsize DC-UNet} & \scriptsize{Only NA} & 0.7633 & 0.6200 & 0.7984 & 0.7638 & 0.6256 & 0.8010 \\
   & \textbf{07:03} & \textbf{0.8962} & \textbf{0.8468} & \textbf{0.8968} & \textbf{0.8982} & \textbf{0.8509} & \textbf{0.8976} \\
\hline

{\scriptsize Half} & \scriptsize{Only NA} & 0.7803 & 0.5127 & 0.8169 & 0.7835 & 0.5170 & 0.8198 \\
{\scriptsize -UNet}  & \textbf{07:03} & \textbf{0.8795} & \textbf{0.8060} & \textbf{0.8918} & \textbf{0.9036} & \textbf{0.8275} & \textbf{0.9090} \\
\hline

{\scriptsize Attention} & \scriptsize{Only NA} & 0.7797 & 0.7377 & 0.8088 & 0.7798 & 0.6564 & 0.8189 \\
    {\scriptsize UNet}  & \textbf{07:03} & \textbf{0.8782} & \textbf{0.8209} & \textbf{0.8976} & \textbf{0.9061} & \textbf{0.8741} & \textbf{0.9123} \\
\hline

{\scriptsize RRDB} & \scriptsize{Only NA} & 0.7872 & 0.6153 & 0.8217 & 0.7892 & 0.6179 & 0.8258 \\
  {\scriptsize UNet}  & \textbf{07:03} & \textbf{0.8702} & \textbf{0.8411} & \textbf{0.8975} & \textbf{0.8941} & \textbf{0.8625} & \textbf{0.9147} \\
\hline

\end{tabular}
\end{table}

\begin{figure}[!htbp]
   \centering
    \includegraphics[width=0.85\linewidth]{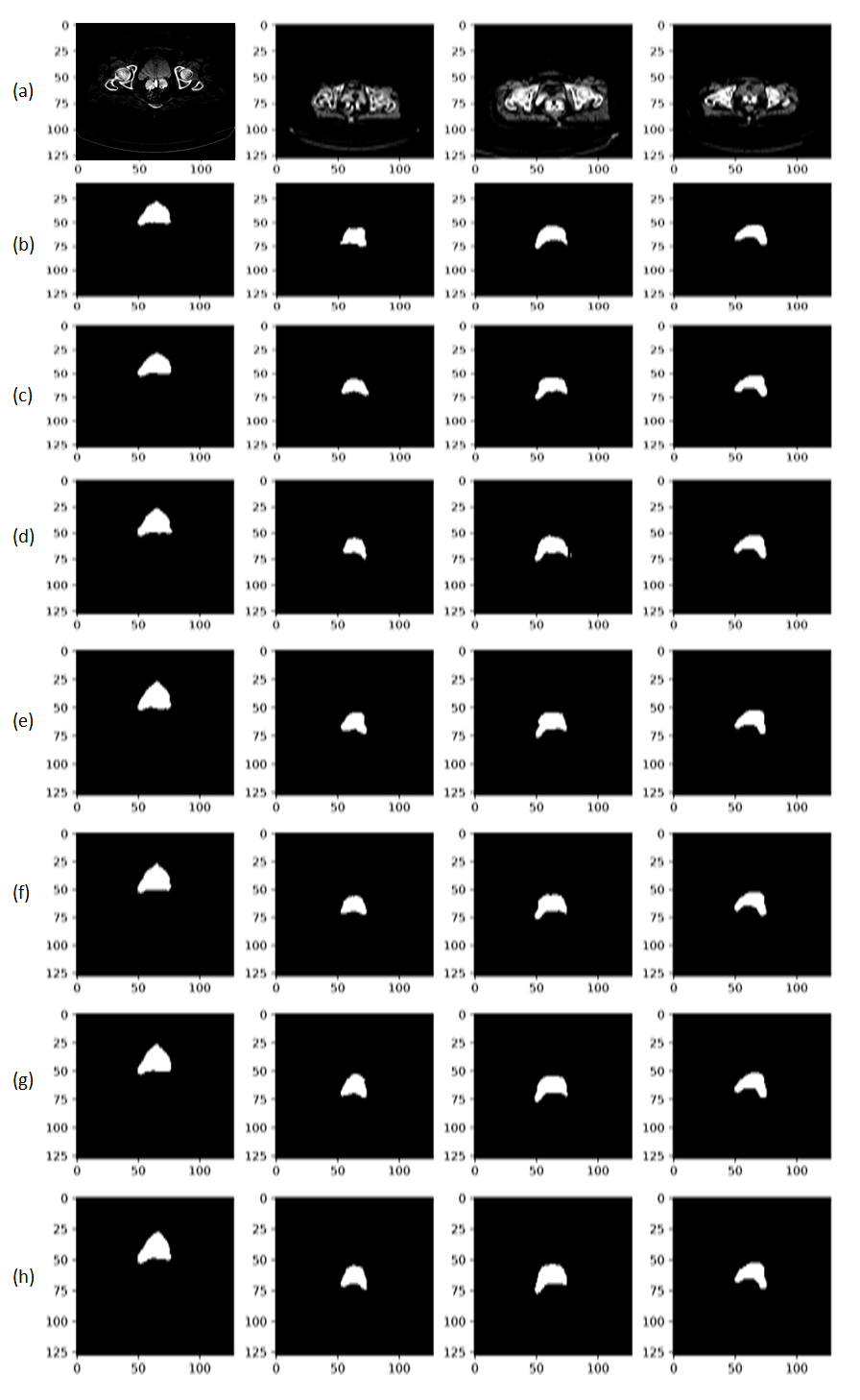}
    \caption{Bladder segmentation in Axial plane for different deep models for \textbf{NA:WA}$=7:3$ proportional data doping. The panels (a) and (b) represent original CT image and corresponding ground truth masks respectively. Panels (c)-(h) represent segmentation results for different models, U-Net, Half-UNet, RRDB U-Net, U-Net++, DC-UNet and Attention U-Net with the dice loss.}
\label{fig:differentimage}
\end{figure}

\subsection{Generalization of Data Doping}
To generalize our findings on data doping, we evaluate this strategy for six different deep learning models. As obtained in the previous section, we consider our training data for this section as \textbf{NA:WA}$=07:03$. A comparative analysis is presented in Table \ref{tab:model_results} considering the six deep learning networks i.e U-Net, Half-UNet, RRDB U-Net, U-Net++, DC-UNet, Attention U-Net under the dice loss. For comparison, we also report results for solely \textbf{NA} training for all these models. A massive average improvement of 19.71\% in U-Net, 9.25\% in U-Net++, 19.52\% in DC-UNet, 17.91\% in Half-UNet, 11\% in Attention U-Net, 16.73\% in RRDB U-Net has been obtained in IoU values, with the axial plane giving the most remarkable results with average increase of 24.45\%. The visual results with optimum mixing for different models have been reported in Fig. \ref{fig:differentimage}. 

\section{Conclusion}
\label{section{Conclusion}}
Performance degradation of deep models due to a covariate shift is a known challenge. However, it is difficult to predict the extent of this degradation as the actual underlying distributions are not known beforehand. This, in turns, make it an open question that whether sample from one distribution might help in learning with the samples from the shifted distribution. We address this critical issue for brachytherapy in cervical cancer patients, where applicator-inserted WA datasets are rare. Due to severe scarcity of WA data, deep models tend to underfit, making it nontrivial to assess whether augmenting training with anatomically similar NA data could yield performance gains while mitigating the covariate shift issue. In this paper, we have shown that, for the downstream task of bladder segmentation in gynecological brachytherapy, to address the scarcity of corresponding samples, the inclusion of \textbf{NA} images in training dataset provides a robust and generalized segmentation model for \textbf{WA} CT images, adapting to anatomical variations induced by the applicator. By mixing both types of data, the model is exposed to a diverse range of bladder shapes and positions, improving its ability to handle clinical variability. This approach enhances the model's performance in real-world settings where \textbf{NA}-data is far more abundant compared to \textbf{WA} scenarios in brachytherapy treatment. While bladder segmentation is a critical task for the patients undergoing such procedures, lack of data impacts the automated segmentation process, which can be augmented effectively with \textbf{NA}-data. 
With only 10-30\% of \textbf{WA} data present in the training set, the segmentation is performed with precise accuracy for a number of standard deep  models. This proposed  strategy of data doping handles the intensity artifacts and structural distortions introduced by the applicator  exceptionally well, through dual-domain learning. The proposed method demonstrates improved Dice Similarity Coefficient (DSC) and Intersection over Union (IOU) scores, indicating better contour overlap and segmentation reliability. This method ultimately supports safer and more accurate brachytherapy treatment planning by ensuring consistent bladder localization, for gynecological cancer, overcoming the critical issue of data scarcity.

\section*{ Declarations}
\subsection*{Conflict of Interests}
The authors have no relevant financial or non-financial interests to disclose.
\subsection*{Funding information}
The authors declare that no funds, grants, or other support were received during the preparation of this manuscript.

\bibliographystyle{unsrt}
\bibliography{bibliography.bib}
\end{document}